\documentclass{article}
\usepackage{spconf,amsmath,graphicx}
\usepackage{float}
\usepackage{enumitem}
\setlist{nosep, leftmargin=14pt}
\usepackage{multirow}
\usepackage{mwe} 
\usepackage{stfloats} 
\usepackage{tabularx}
\usepackage{amsmath,amssymb,graphicx,booktabs,xcolor,cite}

\title{Multimodal Fusion Methods for Whole-Slide Images and Clinical Data in Inflammatory Bowel Disease}

\name{%
    \begin{tabular}{@{}c@{}}
        Ankana Banerjee$^{1}$, 
        Mark van der Giezen$^{1,5,6}$, 
        Ole Gunnar Aasprong$^{2}$, 
        Emiel A. M. Janssen$^{1,2}$ \\
        Øyvind Steinsbø$^{3,7}$, 
        Tore Grimstad$^{3,7}$, 
        Kjersti Engan$^{4}$%
        \thanks{Corresponding author: Ankana Banerjee (ankana.banerjee@uis.no)}
    \end{tabular}%
}

\address{%
    \small
    \begin{tabular}{@{}c@{}}
        $^{1}$Department of Chemistry, Bioscience and Environmental Engineering, University of Stavanger, Norway \\
        Department of Pathology$^{2}$, Dept. of Internal Medicine$^{3}$ (Unit of Gastroenterology), and Research$^{5}$, Stavanger University Hospital, Norway \\
        $^{4}$Department of Electrical Engineering and Computer Science, University of Stavanger, Norway \\
        $^{6}$Natural Resources Institute, University of Greenwich, Chatham, UK \\
        $^{7}$Dept. of Clinical Science, University of Bergen, Bergen, Norway
    \end{tabular}%
}
\begin{document}
%
\maketitle
\begin{abstract}
Multimodal AI fusion of histology and clinical data is an active research area in computational pathology, but fusion methods are rarely compared under identical conditions and across different clinical endpoints. 
We present a modular framework in which slide-level embeddings, obtained by training an attention-based aggregator over patch embeddings from a frozen foundation model, and per-visit clinical embeddings are fed into 15 interchangeable fusion methods with a shared output dimension and prediction head (except decision-level methods).
Using a prospective cohort of newly diagnosed inflammatory bowel disease patients, we predicted STRIDE-II-aligned remission outcomes from baseline or follow-up visits.  
The best fusion method reached macro-F1 of 0.57-0.75, outperforming both single-modality models in four of six scenarios, and differed by outcome and input. Image reliance, the drop in balanced accuracy when the slide embedding is removed, tended to follow the remission type - highest for histological and lowest for endoscopic remission; methods with similar accuracy relied on the slide to very different degrees.

\end{abstract}
\section{Introduction}
\label{sec:intro}

Multimodal learning combining whole-slide histology images (WSIs) and clinical measurements to predict clinical outcomes in IBD is growing in computational pathology \cite{ding2025multimodal, Rieder2026IBDome}, as is fusing many other types of healthcare domain data \cite{demrozi2025multimodal,iacucci2025endo,mossotto2017classification}. 
WSIs capture tissue morphology that clinical records cannot, while clinical records add context the images lack. 
However, fusion methods are typically benchmarked on a single task with a fixed input configuration \cite{Cui2023DeepReview, Steyaert2023MultimodalTumors} and rarely compared for robustness under identical conditions across diverse clinical endpoints.  

STRIDE-II, a treat-to-target strategy for inflammatory bowel disease (IBD), replaces symptom-driven care with objective remission targets - clinical, biochemical, and endoscopic- alongside histological remission as a measure of deeper healing - that resolve on different time scales \cite{Turner2021STRIDE-II:IBDc}. These targets reflect complementary aspects of disease; for example, histology captures microscopic-level inflammation while clinical biomarkers add complementary context to that. Fusing them could therefore yield a better characterization than either alone.  We hypothesize that the best fusion method depends on the prediction scenario rather than being universal. 

\begin{figure*}[t!]
\centering
\includegraphics[width=0.85\textwidth]{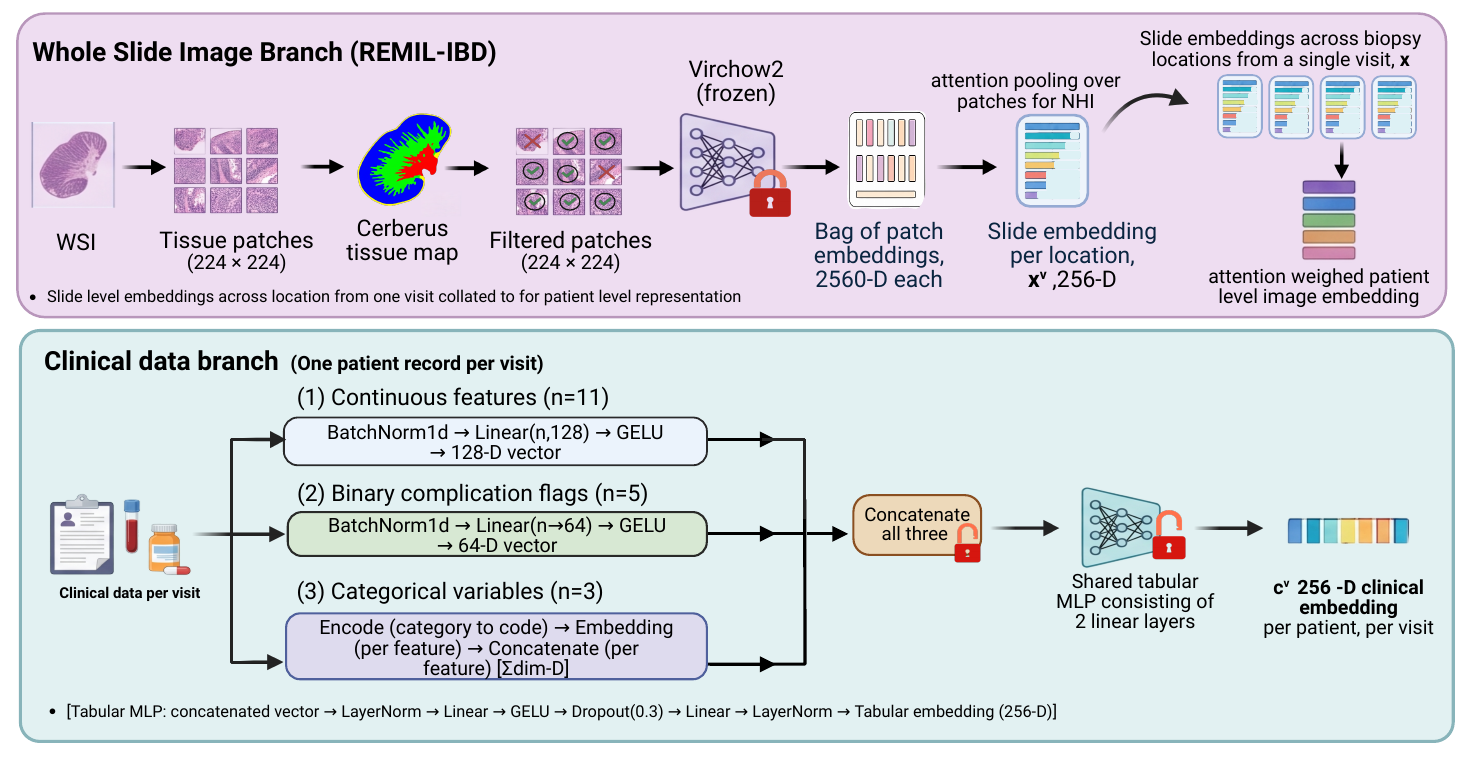}
\caption{Whole Slide image and clinical data embedding generation framework.}
\label{fig:framework}
\end{figure*}

We therefore evaluate 15 interchangeable fusion methods, each with a shared output dimension and prediction head, across STRIDE-II endpoints and input visits (baseline or follow-up visit). The aggregator, fusion method, and prediction head are trained jointly so that performance differences can be attributed primarily to the fusion method.

\section{Data Material}
\label{data}
We used the prospective, single-center SUSI-IBD cohort of newly diagnosed IBD patients (REK 2011/2631; written informed consent), with visits at diagnosis ($V_0$), 3 months ($V_3$), and 11 months ($V_{11}$). Biopsy sampling, scanning, and Nancy Histologic Index (NHI) grading follow REMIL-IBD~\cite{Banerjee2026REMIL-IBD:Disease}. Three STRIDE-II outcomes at $V_{11}$ are: histological (maximum NHI $\leq 1$), endoscopic (MES $\leq 1$ or SES-CD $\leq 2$) and biochemical remission ($\text{CRP}<5~\text{mg/L}$ and faecal calprotectin $<150$~\textmu g/g).
The per-visit clinical input comprised 11 continuous variables (age, BMI, CRP, fecal calprotectin, albumin, hemoglobin, ferritin, transferrin saturation, leukocyte count, plus unified endoscopic and clinical scores derived from subtype-specific MES/PMS and SES-CD/HBI), five binary complication indicators (flare-up, drug side effects, infection, surgery, other complication), three categorical variables (sex, medication and disease subtype). NHI was excluded from clinical input so that histology enters only through the slides.
Patients were included if slides at the input visit and the outcome at $V_{11}$ were available. This gave 250-265 patients for $V_0$ and 241-258 for $V_3$, of whom about 57-74\% were in remission at $V_{11}$. We split data at the patient level using the REMIL-IBD split~\cite{Banerjee2026REMIL-IBD:Disease} to prevent data leakage.

\section{multimodal AI framework}
\label{sec:methods}

\subsection{Whole Slide Image encoder}
The multimodal AI framework is outlined in Figure \ref{fig:framework}. The image encoder is the slide-level encoder from REMIL-IBD \cite{Banerjee2026REMIL-IBD:Disease}, a region-filtered embedding-based multiple instance learning framework that grades histological inflammation from H\&E WSIs of IBD biopsies. WSIs are downsampled (20$\times$), background-filtered, and segmented using Cerberus \cite{Graham2023OneClassification}. Patches were retained if their tissue map contains any abnormal epithelium, inflammatory tissue, or debris, alongside $\geq$25\% normal epithelium or $\geq$15\% stroma. To avoid resizing interpolation, native 224$\times$224 patches are extracted via coordinate back-projection. Patches were encoded by the frozen foundation model Virchow2 (2560-D) \cite{Zimmermann2024Virchow2:Pathology}, pooled by gated attention \cite{Ilse2018Attention-basedLearning} for NHI grade prediction. Optimized using focal ordinal loss \cite{liu2022novel} for 50 epochs with Adam (learning rate $1\times10^{-5}$), selecting the highest validation macro-F1 as the checkpoint. From each WSI, the 256-D output of the gated-attention pooling layer, preceding the classification head, served as the slide embedding.

For patient $i$ and visit $v\in\{V_0,V_3\}$, the frozen slide embeddings $\mathbf{s}_{i,j}^{v}$ of biopsy locations $j$ form a bag. A projection layer shared across visits (linear, ReLU, dropout, LayerNorm) is applied to each slide embedding, and a visit-specific gated-attention network returns their attention-weighted sum as one patient-level 256-D image embedding $\mathbf{x}_i^{v}$ per visit.

\subsection{Clinical Data Encoder}
The clinical encoder has three parallel branches. Continuous ($n_c$ inputs) and binary ($n_b$ inputs) variables are batch-normalized and projected (Linear-GELU) into 128-D and 64-D embeddings, respectively.
Categorical variables are encoded as integer class indexes and passed through dedicated learned embedding layers: sex and disease type (two categories each) are embedded into 4-D vectors and the medication group (eight categories) into a 16-D vector, concatenated into a 24-D representation.
Branch outputs (128-D continuous, 64-D binary, 24-D categorical) were concatenated to preserve modality-specific information and prevent compression loss from forced resizing \cite{Guo2016EntityVariables}. A two-layer MLP with LayerNorm, GELU, and dropout ($p=0.3$) unifies these subspaces into a weighted 256-D per-visit clinical embedding \cite{Gorishniy2021RevisitingData}.

\begin{table*}[h!]
\caption{Fusion methods. $\mathbf{x}\in\mathbb{R}^{256}$ is the pooled image embedding, $\mathbf{c}\in\mathbb{R}^{256}$ the clinical embedding, and $\tilde{\mathbf{s}}_j$ the projected slide embeddings of the input bag(s) before pooling. $\mathbf{z}$ is the fused vector passed to the prediction head (512-D for concat, tabAE and cross-attention, 256-D otherwise); decision-level methods output logits $\boldsymbol{\ell}$ or class probabilities $\mathbf{p}$ per modality. }
\vspace{2pt}
\label{tab:fusion}
\centering\footnotesize
\begin{tabular}{@{}p{1.5cm}p{2.6cm}p{7cm}p{5.5cm}@{}}
\toprule
Type & Method & Description & Equation \\
\midrule
Concatenation & \texttt{concat} & Embeddings are joined side by side. & $\mathbf{z}=[\mathbf{x};\mathbf{c}]\in\mathbb{R}^{512}$ \\
& \texttt{tabAE} & Same as \texttt{concat}, but the clinical encoder $E_\phi$ is first pretrained as an autoencoder, where a decoder $D_\psi$ reconstructs the continuous and ordinal features $\mathbf{r}$ from $\mathbf{c}$. & $\mathbf{z}=[\mathbf{x};\mathbf{c}]$, \; $\mathbf{c}=E_\phi(\cdot)$, \; $\mathcal{L}_{\mathrm{AE}}=\|D_\psi(\mathbf{c})-\mathbf{r}\|_2^2$ \\

& \texttt{early concat} & $\mathbf{c}$ is appended to every slide embedding, projected, then attention-pooled. & $\tilde{\mathbf{s}}'_j=\mathrm{ReLU}(\mathbf{W}[\tilde{\mathbf{s}}_j;\mathbf{c}])$, \; $\mathbf{z}=\mathrm{pool}(\{\tilde{\mathbf{s}}'_j\})$ \\
\midrule

Modulation & \texttt{film} & $\mathbf{c}$ produces a scale $\boldsymbol{\gamma}$ and shift $\boldsymbol{\beta}$ that modulate $\mathbf{x}$ \cite{perez2018film}. & $\mathbf{z}=(\mathbf{1}+\boldsymbol{\gamma}(\mathbf{c}))\odot\mathbf{x}+\boldsymbol{\beta}(\mathbf{c})$ \\
& \texttt{early film} & Same modulation as \texttt{film}, but applied to every slide embedding $\tilde{\mathbf{s}}_j$ before pooling. & $\tilde{\mathbf{s}}'_j=(\mathbf{1}+\boldsymbol{\gamma}(\mathbf{c}))\odot\tilde{\mathbf{s}}_j+\boldsymbol{\beta}(\mathbf{c})$, \; $\mathbf{z}=\mathrm{pool}(\{\tilde{\mathbf{s}}'_j\})$ \\
\midrule

Gating & \texttt{gmu} & Each modality is projected, then the $\tanh$-activated outputs ($\mathbf{h}_x$, $\mathbf{h}_c$) are mixed by a sigmoid gate $\mathbf{g}$ that weights the two vectors \cite{arevalo2017gated}. & $\mathbf{h}_x=\tanh(\mathbf{W}_x\mathbf{x})$, \; $\mathbf{h}_c=\tanh(\mathbf{W}_c\mathbf{c})$, \; $\mathbf{g}=\sigma(\mathbf{W}_g[\mathbf{x};\mathbf{c}])$, \; $\mathbf{z}=\mathbf{g}\odot\mathbf{h}_x+(\mathbf{1}-\mathbf{g})\odot\mathbf{h}_c$ \\
\midrule

Bilinear & \texttt{mcb} & Fixed random count sketches $S(\cdot)$ of the projected embeddings are multiplied elementwise, then normalized (signed square root, $\ell_2$) \cite{fukui2016multimodal}. & $\mathbf{u}=S(\mathbf{W}_x\mathbf{x})\odot S(\mathbf{W}_c\mathbf{c})$, \; $\mathbf{z}=\mathrm{LN}\big(\ell_2(\mathrm{sign}(\mathbf{u})\sqrt{|\mathbf{u}|})\big)$ \\
& \texttt{mcb transformer} & Image, clinical and MCB output $\mathbf{m}$ form three tokens for a one-layer transformer encoder. & $\mathbf{m}=\mathrm{MCB}(\mathbf{x},\mathbf{c})$, \; $\mathbf{T}=\mathrm{Enc}_1([\mathbf{W}_x\mathbf{x};\mathbf{W}_c\mathbf{c};\mathbf{m}])$, \; $\mathbf{z}=\mathrm{GELU}(\mathbf{W}\,\mathrm{LN}[T_1;T_2;\bar T])$ \\
\midrule

Attention & \texttt{temporal concat} & Image and clinical tokens attend to each other in a one-layer transformer encoder \cite{Vaswani2017AttentionNeed}; the two output tokens $T_1$ and $T_2$ are then joined and projected (layernorm, linear layer, GELU). & $[T_1;T_2]=\mathrm{Enc}_1([\mathbf{W}_x\mathbf{x};\mathbf{W}_c\mathbf{c}])$, \; $\mathbf{z}=\mathrm{GELU}(\mathbf{W}\,\mathrm{LN}[T_1;T_2])$ \\

\addlinespace[2pt]
& \texttt{cross-attention} & $\mathbf{c}$ queries the unpooled slide embeddings $\tilde{\mathbf{s}}_j$; the result is joined with $\mathbf{c}$. & $\alpha_j=\mathrm{softmax}_j\!\big(\tfrac{(\mathbf{W}_Q\mathbf{c})^\top\mathbf{W}_K\tilde{\mathbf{s}}_j}{\sqrt d}\big)$, \; $\mathbf{a}=\mathrm{LN}\big(\sum_j\alpha_j\mathbf{W}_V\tilde{\mathbf{s}}_j\big)$, \; $\mathbf{z}=[\mathbf{a};\mathbf{c}]$ \\
\addlinespace[2pt]
& \texttt{cross-attention transformer} & Two-layer bidirectional cross-attention between $\mathbf{x}$ and $\mathbf{c}$, each with a feed-forward blocks \cite{tsai2019multimodal}. & $\mathbf{z}=\mathrm{GELU}(\mathbf{W}\,\mathrm{LN}[\mathbf{x}_L;\mathbf{c}_L])$ \\

\midrule
Decision-level & \texttt{late weighted} & Learned single weight on modality logits. & $\mathbf{z}=\alpha\boldsymbol{\ell}_x+(1-\alpha)\boldsymbol{\ell}_c$, \; $\alpha=\sigma(a)$ \\
& \texttt{product of experts} & Class probabilities multiplied and renormalised \cite{hinton2002training}. & $\mathbf{p}=\dfrac{\mathbf{p}_x\odot\mathbf{p}_c}{\sum_k p_{x,k}\,p_{c,k}}$ \\
& \texttt{decision ensemble} & Class probabilities averaged. & $\mathbf{p}=\tfrac12(\mathbf{p}_x+\mathbf{p}_c)$ \\
& \texttt{confidence weighted} & Each branch is weighted by its maximum class probability (no learned parameters). & $\mathbf{p}=\dfrac{w_x\mathbf{p}_x+w_c\mathbf{p}_c}{w_x+w_c}$, \; $w_m=\max_k p_{m,k}$ \\
\bottomrule
\end{tabular}
\end{table*}

\subsection{Fusion methods}
\begin{figure}[b!]
\centering
\includegraphics[width=\columnwidth]{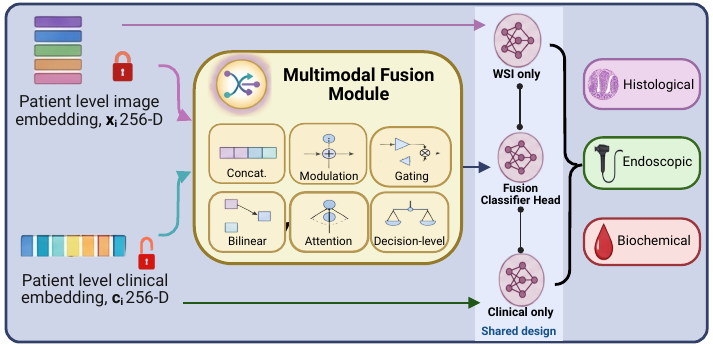}
\caption{Fusion pipeline. Slide image embeddings and clinical embeddings enter one of 15 interchangeable fusion modules.}
\label{fig:fusion}
\end{figure}
The fusion pipeline is depicted in \ref{fig:fusion}. For patient $i$ and input visit(s) $v\in\{V_0,V_3\}$, the fusion modules operate on the following quantities:
\textbf{Projected slide embeddings:} $\tilde{\mathbf{s}}_{i,j}^{v}\in\mathbb{R}^{256}$, the elements $j$ of the visit-$v$ bag after the shared projection layer, before attention pooling into the visit-level embedding $\mathbf{x}_i^{v}$.
\textbf{Image embedding:} $\mathbf{x}_i\in\mathbb{R}^{256}$, obtained by attention-pooling the projected slide embeddings of each visit. 
\textbf{Clinical embedding:} $\mathbf{c}_i\in\mathbb{R}^{256}$, the output of the clinical encoder applied to the clinical measurements of the input visit(s), computed in a single pass.
We compared 15 fusion methods (Table~\ref{tab:fusion}), grouped by the stage at which the modalities interact. Most operate on the pooled embeddings $\mathbf{x}_i$ and $\mathbf{c}_i$; early concatenation, early FiLM, and cross-attention instead act on the projected slide embeddings $\tilde{\mathbf{s}}_{i,j}$, before or in place of attention pooling. Intermediate and early fusion outputs feed a common head (LayerNorm, dropout, Linear$\to$256, GELU, dropout, Linear$\to C$); only its input width differs for 512-D concatenation-type outputs. Late fusion methods bypass it: a linear classifier per modality produces predictions that are combined and trained through the combined output.

\section{Experiments and results}
\label{sec:eval}

We evaluate six combinations of $V_{11}$ outcome (histologic, endoscopic, biochemical remission) and input from $V_0$ or $V_3$, each with WSI-only and clinical-only baselines and 15 fusion methods. Test sets had 26-31 patients, and each model was trained once (seed 42), so intervals reflect test-set sampling, not training variability. The slide embeddings are frozen; all other components were trained end-to-end per task with focal loss ($\gamma=2$, inverse square-root class weights), AdamW (learning rate $10^{-4}$, weight decay $10^{-4}$), cosine annealing to 1\% of the initial rate, gradient clipping at 1.0, 50 epochs, selecting the checkpoint with the highest validation macro-F1.

\begin{figure}[h!]
\includegraphics[width=\columnwidth]{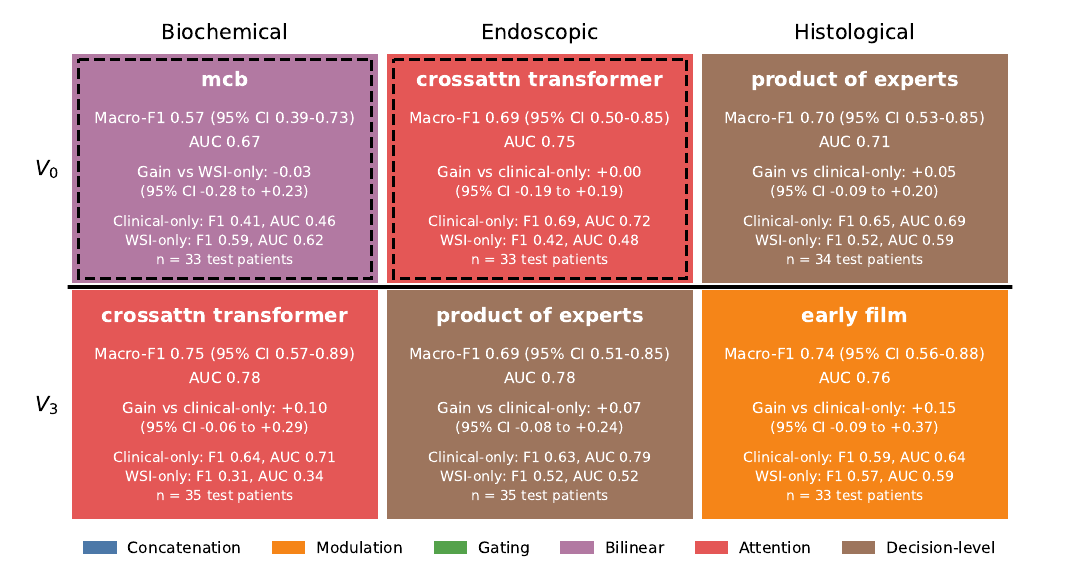}
\caption{Best fusion method in each of the six scenarios.
}
\label{fig:best}
\end{figure}
\begin{figure}[h!]
\includegraphics[width=0.93\columnwidth]{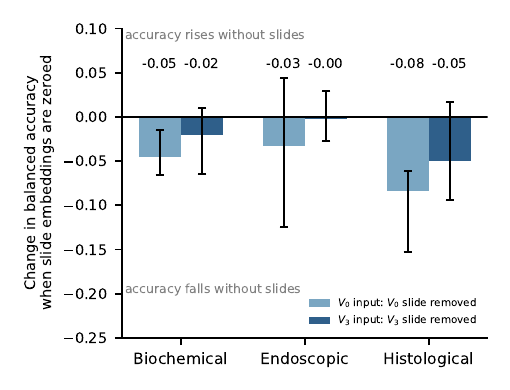}
\caption{Change in balanced accuracy when the slide embedding is replaced by zeros, by outcome and input visit (Absolute cohort). Bars show the median over the fusion methods above chance, and whiskers show the 25th to 75th percentile.}
\label{fig:imageReliance}
\end{figure}

\subsection{Experiment 1 - fusion methods}
The primary metric is test macro-F1, with 95\% percentile intervals from 2000 patient-level bootstrap draws (paired for comparisons). In each scenario, we report the best fusion method and those within 2 and 5 percentage points of its macro-F1. The best fusion method reached a macro-F1 of 0.57-0.75 (Figure~\ref{fig:best}). Fusion performed better than the single-modality model in four scenarios and was comparable to it in two: $V_0$ inputs for biochemical (comparable with WSI-only) and endoscopic remission (same as clinical-only). The clinical-only model was often competitive, so much of the fused performance probably comes from the clinical embedding. The best method exceeded the median method by 0.06-0.15 macro-F1, so fusion's success depends on finding a suitable method rather than combining modalities alone.

The best method depended on both outcome and input. With $V_0$ as input, bilinear (mcb), attention (cross-attention), and decision-level (product of experts) methods performed best for biochemical, endoscopic, and histological remission, respectively. With $V_3$ as input, attention (cross-attention), decision-level (product of experts), and modulation (early film) methods performed best. Method rankings did not carry over between scenarios (mean Spearman $\rho=-0.08$ for macro-F1). In every scenario, at least one other method was within 5\% of the top macro-F1, and up to six in histological remission with $V_0$, from unrelated families (mixing concatenation, attention, bilinear and decision-level families).

\subsection{Experiment 2 - image input reliance}

We define \textit{Image reliance} as the drop in test balanced accuracy when the 256-D slide-embedding is replaced by zero vectors (attention mask zeroed), keeping the trained model unchanged, so that every patient receives the same uninformative image representation. Methods with near or below chance performance (balanced accuracy $\leq 0.5$) are excluded. 
Because the top-ranked fusion method is identified on the test set, we checked if rankings generalize by correlating macro-F1 relative to the scenario median between scenarios (Spearman) against a permutation null (5000 shuffles).

\textit{Image reliance} tended to follow the remission type: it was highest for histological remission with both inputs, but differences between outcomes were not statistically significant (Kruskal-Wallis $p=0.23$ for $V_0$, $p=0.17$ for $V_3$). Removing the slides lowered balanced accuracy, highest in histological and least in endoscopic remission; see Figure \ref{fig:imageReliance}.
The best method in each scenario relied heavily on the WSIs (a drop in raw test balanced accuracy of 0.06-0.23, against scenario medians of 0.00-0.08), whereas 21-50\% of methods had a drop of zero or less. Bilinear and decision-level methods rely on the WSI most (median 0.08 and 0.07) and attention and modulation methods least (0.00 and $-0.01$; $p=0.046$). Reliance rose with a model's gain over chance (Spearman $\rho=0.39$).

\section{Conclusion}

Combining WSIs with clinical data is not a single design decision: how the two are combined, and how much the model uses the image, varies with the remission endpoint and the information available. Studies of multimodal fusion should therefore evaluate several fusion methods across endpoints and report image reliance next to accuracy. Larger cohorts, repeated training, multi-timepoint inputs and external validation are needed to show which approaches are robust.

\section{Compliance with Ethical Standards}
This study was performed in line with the principles of the Declaration of Helsinki. We used samples from the SUSI-IBD cohort (NCT01551563), approved by the Norwegian Regional Ethics Committee (REK 2011/2631), and obtained written informed consent from all participants.

\section{Acknowledgments}
This work was supported by the Norwegian Research Council (project no. 324516). The authors thank the nurses and physicians of the Gastroenterology Unit, the Pathology Unit, and the Research Department at Stavanger University Hospital, and the participating patients. We thank the University of Stavanger's Department of Chemistry, Bioscience, and Environmental Engineering and the Department of Electrical Engineering and Computer Science for their support. Tore Grimstad has served as a speaker for Ferring Pharmaceuticals, as a consultant for Takeda AS, Galapagos, Johnson \& Johnson, and Emiel Janssen, and has received unrestricted research grants from AbbVie AS and Tillotts Pharma AB. Writefull, Claude, and ChatGPT were used to improve grammar and language and for code debugging; the authors reviewed all content and take full responsibility for it.

\bibliographystyle{IEEEbib}
\footnotesize

\end{document}